%% file: supp_v3.tex
\documentclass[10pt,twocolumn,letterpaper]{article}

\usepackage{cvpr}              
\input{math_commands.tex}

\usepackage{graphicx}
\usepackage{amsmath}
\usepackage{amssymb}
\usepackage{booktabs}
\usepackage{multirow}
\usepackage{xcolor}
\usepackage{paralist}
\usepackage[noline,ruled]{algorithm2e}
\usepackage[accsupp]{axessibility}
\newcommand{\increase}[1]{\textcolor{green}{#1}}
\newcommand{\decrease}[1]{\textcolor{red}{#1}}

\usepackage{float}

\newcommand{\ziquan}[1]{\textcolor{black}{#1}}
\newcommand{\abc}[1]{\textcolor{black}{#1}}

\newcommand{\CUT}[1]{}
\usepackage[pagebackref,breaklinks,colorlinks]{hyperref}
\usepackage[misc,geometry]{ifsym}

\usepackage[capitalize]{cleveref}
\crefname{section}{Sec.}{Secs.}
\Crefname{section}{Section}{Sections}
\Crefname{table}{Table}{Tables}
\crefname{table}{Tab.}{Tabs.}

\def\cvprPaperID{166} 
\def\confName{CVPR}
\def\confYear{2023}

\begin{document}

\title{TWINS: A Fine-Tuning Framework for Improved Transferability of Adversarial Robustness and Generalization}

\author{Ziquan Liu$^{1}$, Yi Xu$^{2}$\thanks{Corresponding author}, Xiangyang Ji$^3$ and Antoni B. Chan$^{1*}$\\
$^1$Department of Computer Science, City University of Hong Kong\\
$^2$School of Artificial Intelligence, Dalian University of Technology\\
$^3$Department of Automation, Tsinghua University\\
{\tt\small ziquanliu.cs@gmail.com, yxu@dlut.edu.cn, xyji@tsinghua.edu.cn, abchan@cityu.edu.hk}
}
\maketitle

\begin{abstract}
Recent years have seen the ever-increasing importance of pre-trained models and their downstream training in deep learning research and applications. At the same time, the defense for adversarial examples has been mainly investigated in the context of training from random initialization on simple classification tasks. To better exploit the potential of pre-trained models in adversarial robustness, this paper focuses on the fine-tuning of an adversarially pre-trained model in various classification tasks. Existing research has shown that since the robust pre-trained model has already learned a robust feature extractor, the crucial question is how to maintain the robustness in the pre-trained model when learning the downstream task. We study the model-based and data-based approaches for this goal and find that the two common approaches cannot achieve the objective of improving both generalization and adversarial robustness. Thus, we propose a novel statistics-based approach, \textbf{T}wo-\textbf{WI}ng \textbf{N}ormli\textbf{S}ation (TWINS) fine-tuning framework, which consists of two neural networks where one of them keeps the population means and variances of pre-training data in the batch normalization layers. Besides the robust information transfer, TWINS increases the effective learning rate without hurting the training stability since the relationship between a weight norm and its gradient norm in standard batch normalization layer is broken, resulting in a faster escape from the sub-optimal initialization and alleviating the robust overfitting. Finally, TWINS is shown to be effective on a wide range of image classification datasets in terms of both generalization and robustness. Our code is available at \url{https://github.com/ziquanliu/CVPR2023-TWINS}. 
\end{abstract}
\begin{figure}[t]
    \centering
    \includegraphics[width=1.0\linewidth]{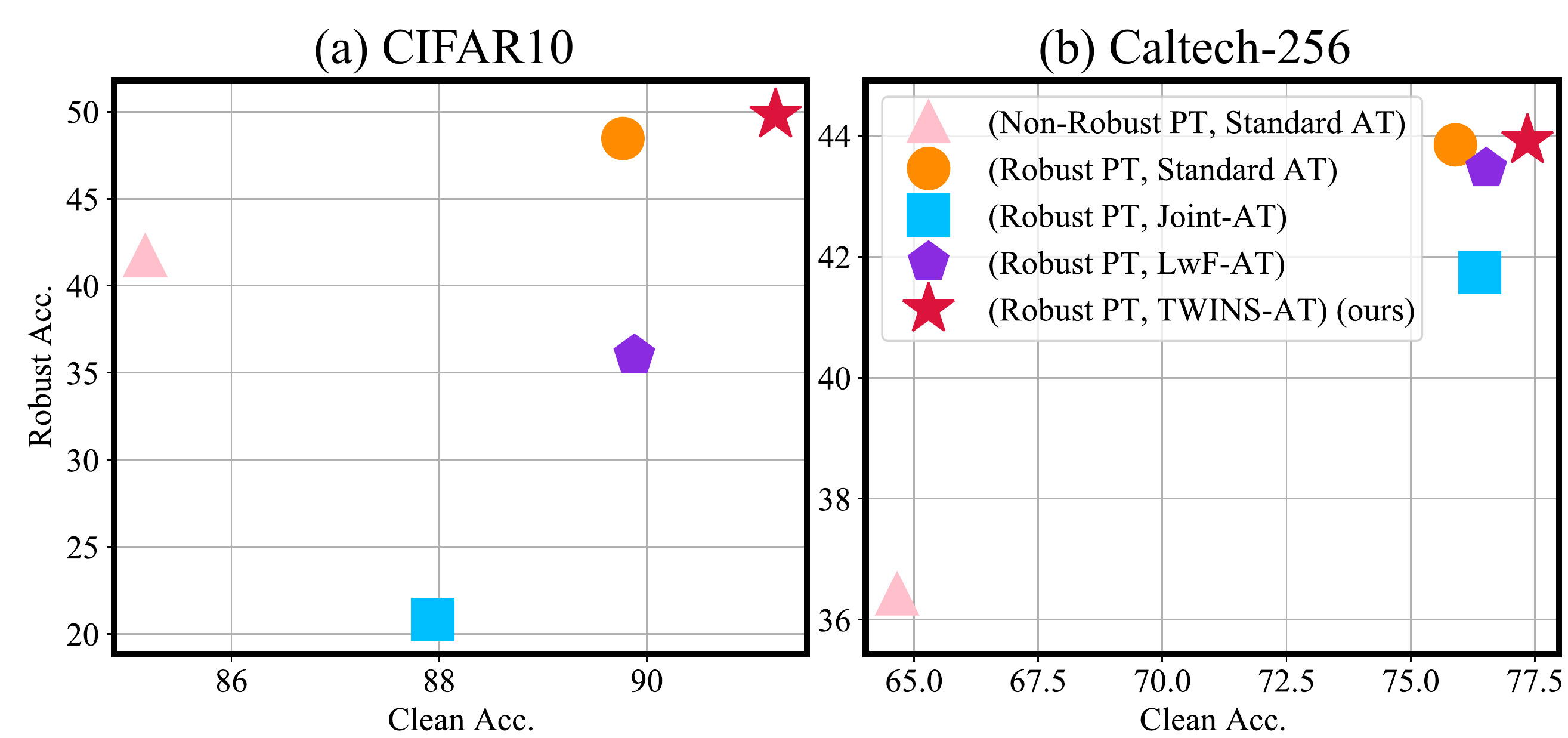}
    \vspace{-0.6cm}
    \caption{The performance of fine-tuning robust and non-robust large-scale pre-trained (PT) ResNet50 \cite{he2016deep,salman2020adversarially} on CIFAR10 \cite{krizhevsky2009learning} and Caltech-256 \cite{griffin2007caltech}. We compare standard adversarial training (AT), Learning without Forgetting (LwF) (\emph{model approach}) \cite{li2017learning}, joint fine-tuning with UOT data selection (\emph{data approach}) \cite{liu2022improved} and our TWINS fine-tuning. The robust accuracy is evaluated using $l_{\infty}$ norm bounded AutoAttack \cite{croce2020reliable} with $\epsilon=8/255$. On CIFAR10, the data-based and model-based approach fail to improve clean and robust accuracy. On Caltech, both approaches improve the clean accuracy but hurt the robust accuracy. Our TWINS fine-tuning improves the clean and robust performance on both datasets. The pink triangle denotes the performance of standard AT with the non-robust pre-trained ResNet50, which drops considerably compared with fine-tuning starting from the robust pre-trained model.
    }    \label{fig:compare_data_and_model_approach}
\end{figure}
\section{Introduction}
\label{sec:intro}
\begin{figure*}[t]
    \centering
    \includegraphics[width=1.0\linewidth]{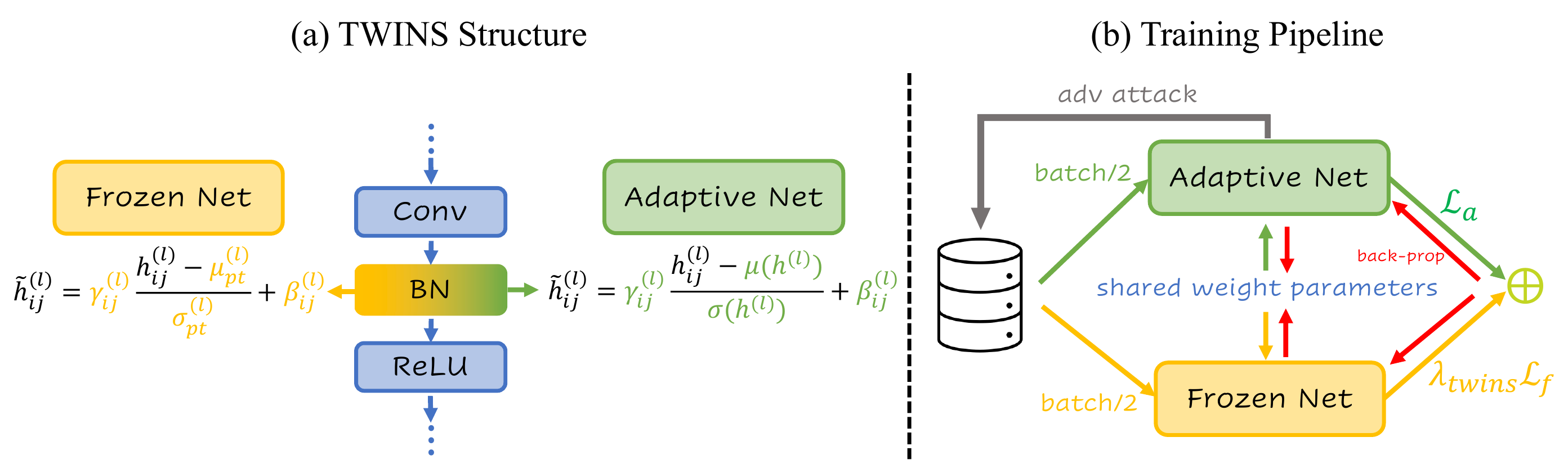}
    \vspace{-0.85cm}
    \caption{The TWINS structure and training pipeline. \textbf{(a)} The Frozen Net and Adaptive Net have the same structure and share the weight parameters, except for batch normalization (BN) layers. The Frozen Net uses pre-trained means and standard deviations (STD) in the normalization layer, while Adaptive Net uses the mean and STD computed from the current batch as in standard BN. \textbf{(b)} In each step of mini-batch stochastic gradient descent (SGD), we split the batch of adversarial examples, generated from attacking the Adaptive Net, into two sub-batches and feed them to the Adaptive Net and Frozen Net respectively. The loss of two networks are combined and back-propagated to their shared parameters to train the network. In the \emph{inference stage}, 
    only the Adaptive Net is used.
    }
    \label{fig:pipeline}
\end{figure*}
The adversarial vulnerability of deep neural networks (DNNs) \cite{szegedy2014intriguing} is one of the major obstacles for their wide applications in safety-critical scenarios such as self-driving cars \cite{feng2021intelligent} and medical diagnosis \cite{finlayson2019adversarial}. Thus, addressing this issue has been one focus of deep learning research in the past eight years. Existing works have proposed to improve adversarial robustness from different perspectives, including data augmentation \cite{madry2018towards,rebuffi2021data,gowal2021improving,sehwag2021robust}, regularization \cite{li2021towards,qin2019adversarial,liu2020improve,liu2022boosting} and neural architecture \cite{huang2021exploring,guo2020meets}. 
However, most of existing works investigate the problem under the assumption that the training data is sufficient enough, and training from scratch gives a satisfactory performance, which is not realistic in the real world. There are a large number of computer vision tasks where training from scratch \abc{is inferior to} 
training from pre-trained weights, such as fine-grained image classification (e.g., Caltech-UCSD Birds-200-2011 or CUB200 \cite{WahCUB_200_2011}), object detection \cite{liu2020deep} and semantic segmentation \cite{minaee2021image}.

On the other hand, pre-trained models have been considered as the foundation models in deep learning \cite{bommasani2021opportunities} as a result of their strong performance and wide employment in computer vision \cite{dosovitskiy2021an,liu2021swin,he2022masked,he2020momentum}, as well as natural language processing \cite{devlin2018bert,brown2020language,raffel2020exploring}. Thus, how to better use the pre-trained model in downstream has emerged as a major research topic in many vision and language tasks, such as image classification under distribution shifts \cite{liu2022empirical,yu2021empirical}, object detection \cite{li2022exploring} and semantic segmentation \cite{li2022mask,jain2021semask}. There are a few papers that investigate the pre-trained model's robustness in target tasks \cite{shafahi2020adversarially,djolonga2021robustness,jiang2020robust,chen2020adversarial,chen2021cartl,jeddi2020simple,yamada2022does}. \cite{shafahi2020adversarially,chen2021cartl} mainly considers the transfer between small-scale datasets (e.g., CIFAR100 to CIFAR10), while \cite{chen2020adversarial,jiang2020robust} use adversarial robust pre-training and fine-tuning on the same dataset, without considering a large-scale and general pre-trained model.  Finally, \cite{djolonga2021robustness,yamada2022does} investigate different kinds of robustness to corruption or out-of-distribution samples, and are not devoted to adversarial robustness. 

\abc{In this paper, we consider how to transfer the adversarial robustness of a large-scale robust pre-trained model (e.g., a ResNet50 pre-trained on ImageNet \cite{deng2009imagenet} with adversarial training) on various downstream classification tasks when fine-tuning with adversarial training.}
\ziquan{This problem setting is becoming more important as the standard pre-trained models do not learn robust representations from the pre-training data and are substantially weaker than the robust pre-trained counterparts in some challenging downstream tasks, e.g., fine-grained classification as shown in our experiment. Meanwhile, more large-scale robust pre-trained models are released (e.g., ResNet \cite{salman2020adversarially} and ViT \cite{bai2021transformers}), which makes the robust pre-trained models more accessible.}
\abc{However, naively applying adversarial training to fine-tune from the robustly pre-trained model will lead to suboptimal robustness, }\ziquan{since the robust representations learned by the robust pre-trained model are not fully utilized. For example, \cite{shafahi2020adversarially} suggests that the robustness from a pre-trained model needs to be explicitly maintained for its better transfer to the downstream.}

Following the idea that the key to improving the transferability of robustness is to maintain the robustness of the pre-training stage during fine-tuning \cite{shafahi2020adversarially}, \ziquan{we first evaluate the data-based and model-based approach on two representative datasets, CIFAR10 and Caltech-256. The data-based approach uses pre-trained data in the fine-tuning and keeps their performance under adversarial attack, while the model-based approach regularizes the distance of features of the fine-tuned and pre-trained model.
Our experiment shows that both methods fail to improve the robustness and generalization (Fig.~\ref{fig:compare_data_and_model_approach}), since the two methods are too aggressive in retaining the robustness and hurt the learning in downstream.
Thus, we propose a subtle approach that keeps the batch-norm (BN) statistics of pre-training for preserving the robustness, which we call {\bf T}wo-{\bf WI}ng {\bf N}ormali{\bf S}ation (TWINS) fine-tuning. TWINS has two neural networks with fixed and adaptive BN layers respectively, where the fixed BN layers use the population means and STDs of pre-training for normalization, while the adaptive BN layers use the standard BN normalization. Our experiment first demonstrates the importance of pre-trained BN statistics in the robust fine-tuning and then finds the benefit of TWINS in adversarial training dynamics. As the relationship between weight norm and its gradient norm no longer holds in TWINS, it is able to increase the gradient magnitude without increasing the gradient variance. At the initial training stage, TWINS has a faster escaping speed from the sub-optimal initialization than vanilla adversarial training \cite{liu2020loss}. At the final training stage, the gradient of TWINS is more stable than adversarial training, which alleviates the robust overfitting effect \cite{sehwag2021robust}. In summary, the contributions of our paper are as follows:} 

\begin{compactenum}
    \item We focus on the fine-tuning of \emph{large-scale} robust pre-trained models as a result of their potential importance in various downstream tasks. We evaluate 
    current  approaches to retain 
    the pre-training robustness in fine-tuning, and show that they 
    cannot substantially improve the robustness.
    \item We propose TWINS, a statistics-based approach for better transferability of robustness and generalization from the pre-training domain to the target domain. TWINS has two benefits: a) it keeps the robust statistics for downstream tasks, thus helps the transfer the robustness to downstream tasks and b) it enlarges the gradient magnitude without increasing gradient variance, thus helps the model escape from the initialization faster and mitigates robust overfitting. The mechanisms of these two benefits are validated by our empirical study. 
    \item The effectiveness of TWINS is corroborated on five downstream datasets by comparing with two popular adversarial training baselines, adversarial training (AT) \cite{madry2018towards} and TRADES \cite{zhang2019theoretically}. On average, TWINS improves the clean and robust accuracy by 2.18\% and 1.21\% compared with AT, and by 1.46\% and 0.69\% compared with TRADES. The experiment shows the strong potential of robust pre-trained models in boosting downstream's robustness and generalization when using more effective  fine-tuning methods.
\end{compactenum}

\CUT{
The structure of this paper is as follows. Section \ref{sec:related} discusses related works and we present the result of data and model approach in Section \ref{sec:model_and_data}. We propose TWINS in Section \ref{sec:twins} and analyze their benefit in robust information retaining and training dynamics. Section \ref{sec:experiment} shows our empirical study on a wide range of downstream datasets and demonstrates the effectiveness of TWINS over popular adversarial training methods and we conclude our paper in Section \ref{sec:conclusion}.
}

\section{Related Work}
\label{sec:related}
\noindent\textbf{Adversarial defense.} There are several major approaches to improving the adversarial robustness of DNNs. The training of DNNs can be regularized to induce biases that are beneficial to adversarial robustness, such as locally linear regularization \cite{qin2019adversarial}, margin maximization \cite{liu2022boosting,ding2019mma} and Jacobian regularization \cite{jakubovitz2018improving}. The most commonly used adversarial defense is adversarial training (AT) \cite{madry2018towards}, which directly trains the DNN on adversarial examples generated from PGD attack. Later, TRADES \cite{zhang2019theoretically} is proposed to add a KL regularization to AT and achieves stronger adversarial robustness. Our paper proposes TWINS to improve adversarial training in the fine-tuning stage when the initial model is adversarially pre-trained. We compare TWINS-AT and TWINS-TRADES with vanilla AT and TRADES in our experiment and show the strong effectiveness of TWINS in the robust fine-tuning setting.

\noindent\textbf{Fine-tuning for downstream robustness.} Several aspects of robustness in pre-training and fine-tuning have been studied in existing works.
Adversarial contrastive learning \cite{jiang2020robust,chen2020adversarial} is proposed to pre-train on a dataset with contrastive learning and then fine-tune on the \emph{same} dataset, without considering the transferability of robustness from a large-scale pre-trained model to a \emph{different} downstream task. In contrast, our paper investigates a more general problem, where \abc{task-specific} pre-training is not needed for a new task as we use one robust large-scale pre-trained model \abc{trained on ImageNet.}
\cite{salman2020adversarially} considers the robust pre-training on the large-scale ImageNet and its transfer to downstream tasks, but focuses on the performance on clean instead of adversarial images. The Learning-without-Forgetting (LwF) \cite{li2017learning} approach for retaining robustness is shown to be effective in the \emph{small-scale} transfer experiment \cite{shafahi2020adversarially}, but is not effective in our experiment setting \abc{of transfer of large-scale models}.
\ziquan{\cite{jeddi2020simple} proposes a learning rate schedule to improve the adversarial robustness of fine-tuned models, and \cite{dong2021should} proposes robust informative fine-tuning for pre-trained language models to robustly keep pre-training information in downstream. The difference between \cite{jeddi2020simple,dong2021should} and our work is that they assume a standard pre-trained model instead of the adversarial pre-trained model. \cite{djolonga2021robustness,yamada2022does} investigate the performance of pre-trained models in downstream tasks, but the focus is the robustness to out-of-distribution samples instead of adversarial perturbations.}


\noindent\textbf{Batch normalization.} 
\ziquan{There are existing papers proposing the two-branch BN structure for  different purposes with different technical details.}
\cite{sitawarin2021improving} proposes dual normalization for a better trade-off between accuracy and robustness, where the normalization is a weighted sum of normalized clean and adversarial input.  \cite{xie2020adversarial} proposes a similar two-branch BN structure, where one branch is for adversarial examples and the other is for clean examples. The major difference between our work and \cite{sitawarin2021improving,xie2020adversarial} is that both BN branches in TWINS are for adversarial examples and \abc{one branch (Frozen Net)} has fixed BN statistics from pre-training so as to better maintain the pre-trained robustness, 
whereas \cite{sitawarin2021improving,xie2020adversarial} uses clean examples in BN and aims to improve the accuracy for clean images.   

\section{The Model-based and Data-based Approach to Retaining Adversarial Robustness}
\label{sec:model_and_data}
This section introduces the two common approaches for keeping adversarial robustness of pre-training in downstream, \abc{model-based and data-based approaches.} Denote the feature vector output of a neural network as $g_{\vtheta}(\vx)$, the training sample in the downstream task as $\{(\vx_i,\vy_i)\}_{i=1}^N\sim\sP$, and the loss function as $\gL(\vw^Tg_{\vtheta}(\vx)+b,\vy)$, where $(\vw,b)$ are the parameters of last classification layer. We assume that the pre-trained model is trained on adversarial examples generated from the PGD attack \cite{madry2018towards}, where the $l_{\infty}$ norm of the adversarial attack is bounded by $\epsilon$, and during fine-tuning we use adversarial training with 
\abc{the same PGD attack} to obtain adversarial robustness in downstream tasks. 
In short, we consider the robust pre-training and robust fine-tuning setting in this paper, if not specified otherwise.

\paragraph{Model-based approaches.}
We first introduce the model-based approach, \abc{which keeps the pre-trained model $\vtheta_{pt}$ during fine-tuning so as to maintain its robustness.}
The objective function is
\small
\begin{align}
\sum_{(\vx_n,\vy_n)\sim\sP}\gL(\vw^Tg_{\vtheta}(\tilde\vx_n)+b,\vy_n)+\lambda_{LwF}\|g_{\vtheta_{pt}}(\tilde\vx_n)-g_{\vtheta}(\tilde\vx_n)\|_2,
\end{align}
\normalsize
where the adversarial example generated from $\vx_n$ is denoted as $\tilde\vx_n$. \abc{The regularization term of the loss aims} 
to minimize the distance between the features from the pre-trained and the fine-tuned models, which is expected to maintain the robustness of the pre-trained model.
This approach is originally proposed in \cite{li2017learning} to prevent the catastrophic forgetting in continual learning, and is used in \cite{shafahi2020adversarially} to preserve adversarial robustness in transfer learning. Note that \cite{shafahi2020adversarially} uses the LwF method in \emph{standard} fine-tuning instead of robust fine-tuning as in our paper.

\paragraph{Data-based approaches.}
The objective function of the data-based approach is
\small
\begin{align}
\sum_{(\vx_n,\vy_n)\sim\sP}&\gL(\vw^Tg_{\vtheta}(\tilde\vx_n)+b,\vy_n)\nonumber\\
+&\lambda_{UOT}\sum_{(\vx_m,\vy_m)\sim\sQ}\gL(\vw_q^Tg_{\vtheta}(\tilde\vx_m)+b_q,\vy_m),
\end{align}
\normalsize
where $\sP$ and $\sQ$ are data distribution of the target and the pre-training tasks, $\vw_q, b_q$ are the classification layer for pre-trained data from $\sQ$.
This method regularizes the current fine-tuned model feature extractor so that its prediction is still robust on the pre-training data. The joint training method is proposed in \cite{liu2022improved} to improve the performance of fine-tuning in downstream tasks where the training data is not sufficient. 

Next, we test the performance of these two approaches on two standard image classification datasets, CIFAR10 \cite{krizhevsky2009learning} and Caltech-256 \cite{griffin2007caltech}, with results shown in Figure~\ref{fig:compare_data_and_model_approach}.  We use a grid search for learning rate and $\lambda_{LwF}$ ($\lambda_{UOT}$) and report the result of the model with the best robust accuracy. See Section \ref{sec:experiment} and the supplemental material for the experiment setting. On CIFAR10, both approaches fail to improve either clean or robust accuracy; on Caltech-256, the two approaches improve the clean accuracy by a small margin but deteriorate the robustness. One reason why the model- and data-based approaches fail is that the regularization term might be too strong, thus hurting the learning in the downstream.
\begin{figure}[t]
    \centering
    \includegraphics[width=1.0\linewidth]{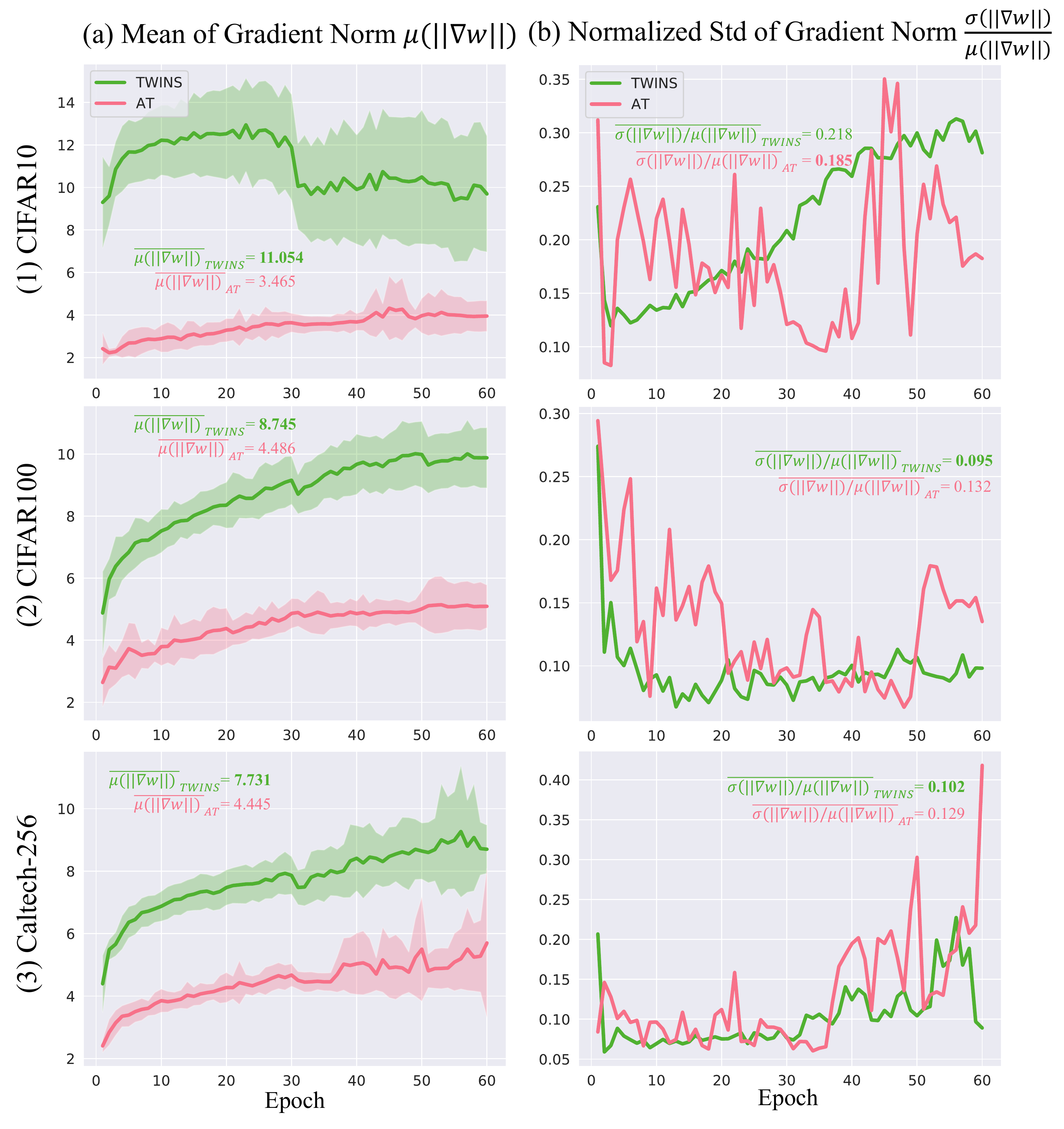}
    \vspace{-0.9cm}
    \caption{The mean and normalized STD of gradient norms in AT and TWINS-AT on CIFAR10, CIFAR100 and Caltech-256. The averaged $\mu(\|\nabla \vw\|)$ and $\sigma(\|\nabla \vw\|)/\mu(\|\nabla \vw\|)$ over epochs are shown in each plot. \ziquan{The gradient magnitudes of TWINS-AT are substantially larger than those of AT, while the \abc{Normalized STDs} of gradient norm in TWINS-AT are not obviously increased (CIFAR10) or even decreased (CIFAR100 and Caltech-256) compared with AT. This property leads to the faster escaping speed of TWINS-AT from the initial sub-optimum compared with AT (Fig.~\ref{fig:weight_dist_compare}), and reduced robust overfitting (Tab.~\ref{tab:robust_overfitting}).}
    }
    \vspace{-0.7cm}
    \label{fig:grad_mean_and_std}
\end{figure}

\section{TWINS Fine-Tuning}
\label{sec:twins}
The previous section demonstrates that both data- and model-based approaches cannot substantially improve the adversarial robustness in the downstream task.
 Thus we propose the TWINS for the better fine-tuning of robust pre-trained models for downstream adversarial robustness.

\subsection{Proposed Method}
Though BN layers contain only a few parameters compared to convolution and fully connected layers, they play an important role in the good performance of DNNs. \cite{frankle2021training} shows that even if we only train the BN layers, the performance of a DNN is already quite impressive. \cite{noguchi2019image} finds that only training the parameters in BN layers in an image generator is effective for small datasets. \cite{li2016revisiting} proposes adaptive BN for domain adaptation, which updates the BN statistics with data from a target domain. These works motivate us to propose a statistics-based approach for retaining pre-training information in the target task. 

Typical BN layers track the mean and STD of the training set and save them for the inference stage.  As this distribution information for each layer might be helpful for downstream robustness, we propose the TWINS robust fine-tuning, 
\abc{which maintains two networks, \emph{Frozen Net} that  uses the BN statistics from the robust pre-trained model, and \emph{Adaptive Net} that learns its BN statistics from the downstream task.}
%
Instead of using two independent networks for the Frozen and Adaptive Net, we let the two networks share weight parameters, excluding the BN layers, to save the model size and inference time. At initialization, both networks and their BN statistics are initialized by the robust pre-trained models. During training, \ziquan{the Frozen Net uses the population means and STDs of pre-training data computed in the pre-training stage in the normalization operation}, while the Adaptive Net uses the current batch's mean and STD in the normalization and updates its running mean and STD with the target training data. Fig.~\ref{fig:pipeline} shows the general pipeline of TWINS training and the network structure. 

\begin{figure}[t]
    \centering
    \includegraphics[width=1.05\linewidth]{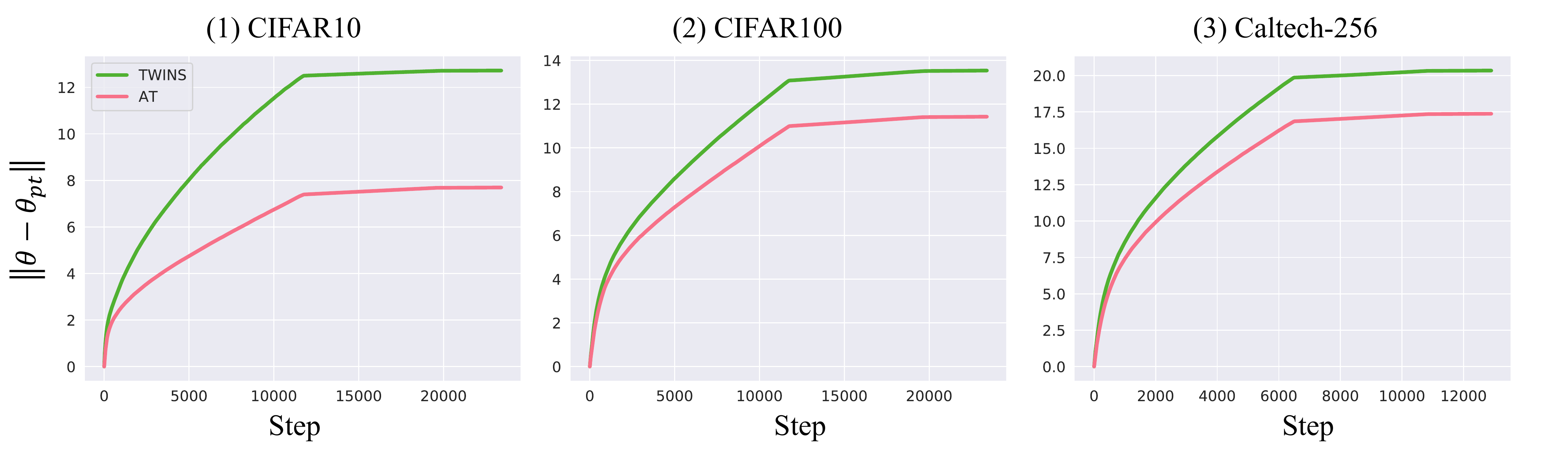}
    \vspace{-0.8cm}
    \caption{The distance between the current step's weight $\vtheta$ and the initialization $\vtheta_{pt}$. On the three datasets, TWINS-AT has a faster escaping speed from the sub-optimal initial model than AT, which is due to TWIN-AT's larger gradient norm as shown in Figure~\ref{fig:grad_mean_and_std}. }
    \vspace{-0.3cm}
    \label{fig:weight_dist_compare}
\end{figure}

The training objective of TWINS with adversarial training (TWINS--AT) in one mini-batch is:
\small
\begin{align}
    &\sum_{j=1}^{B/2}\gL(\vw^T g_{\vtheta_{a}}(\tilde{\vx}_j^{(a)})+b,\vy_j) +\\
    & \quad \lambda_{twins}\sum_{i=B/2+1}^{B}\gL(\vw^T g_{\vtheta_{f}}(\tilde{\vx}_i^{(a)})+b,\vy_i), 
\end{align}
\normalsize
where $\vtheta_f$ and $\vtheta_a$ denote the Frozen and Adaptive Nets respectively, and $\tilde{\vx}^{(a)}$ is the adversarial image for $\vx$ when attacking the Adaptive Net $\vtheta_{a}$. We split the batch into two different sub-batches to avoid doubled batch sizes in TWINS training. Since the Frozen and Adaptive Nets share weight parameters, the number of parameters in TWINS is only increased by a very small amount (i.e., BN parameters). Thus, TWINS-AT only has a negligible cost in terms of memory and training time compared with vanilla AT. The TWINS structure can also be used in the TRADES in a similar way. Similar to \cite{li2016revisiting}, we can use the target training set to update the BN statistics so that they are more relevant to the downstream task. We call this procedure warmup in TWINS fine-tuning. The pseudo codes of TWINS-AT and TWINS-TRADES are given in Alg.~1 of the supplemental.

\subsection{The mechanism of TWINS}
The first benefit of TWINS is mentioned in the motivation of TWINS, i.e., the BN statistics have robustness information in the pre-training domain that can be leveraged by robust fine-tuning to improve the downstream robustness. This argument is validated by our ablation study in Section~\ref{sec:experiment}, where we initialize the means and STDs with 1.0 and 0.0 for Frozen Net instead of the pre-trained means and STDs and check the accuracy and robustness. Figure~\ref{fig:ablation_frozen_bn_no_pretrain} shows that the TWINS with (1,0) initialization cannot match the performance with TWINS with pre-trained statistics, indicating that the robustness information in pre-training is essential to the effectiveness of TWINS. 

It is intriguing that even with the (1,0) initialization, TWINS still outperforms the AT baseline in terms of robustness or accuracy on some datasets, e.g., Stanford Dogs and CIFAR100. This suggests that besides retaining the pre-training information, TWINS provides some other benefits during robust fine-tuning. By analyzing the gradient of TWINS, we find that TWINS implicitly increases the effective learning rate without increasing the oscillation and empirically validate this finding.  We detail this analysis next.

\begin{table}[t]
\begin{center}
\scriptsize
\begin{tabular}{|c|l|l|l|l|l|l|}
\hline
 Method& Rob. Acc.
 &
  C10 &
  C100 &
  Caltech &
  CUB &
  Dogs \\ \hline
\multirow{3}{*}{AT}       & Best $\uparrow$  & 51.84 & 31.38 & 49.09 & 27.08  & 21.19 \\  
                          & Final $\uparrow$ & 49.41 & 28.52 & 48.37 & 26.60   & 19.80 \\ 
                          & Gap $\downarrow$   & 2.43  & 2.86  & 0.73  & 0.48   & 1.39  \\ \hline
\multirow{3}{*}{TWINS-AT} & Best $\uparrow$  & 53.23 & 31.60  & 48.80  & 29.24 & 20.89 \\ 
                          & Final $\uparrow$ & 52.40  & 31.08 & 48.40  & 29.24 & 20.58 \\  
                          & Gap $\downarrow$   & \textbf{0.83}  & \textbf{0.52}  & \textbf{0.40}   & \textbf{0.00}      & \textbf{0.29}   \\ \hline
\end{tabular}
\vspace{-0.3cm}
\caption{The robust accuracy drop of AT and TWINS-AT, where the adversarial attack is PGD10. TWINS-AT has smaller accuracy drop compared with AT, indicating that the TWINS-AT is less prone to robust overfitting as a result of reduced variance of gradient norms and stable training as shown in Fig.~\ref{fig:grad_mean_and_std}.}
\vspace{-0.7cm}
\label{tab:robust_overfitting}
\end{center}
\end{table}
\noindent\textbf{Effective learning rate}
We first write the gradient of a weight vector for TWINS training. Consider the $l$-th layer's weight $\vw^{(l)}_j$ and its output after BN layer
\begin{align}
    \tilde h_{ij}^{(l)}&=\frac{\hat h_{ij}^{(l)}-\frac{2}{B}\sum_{k=1}^{B/2}\hat h_{kj}^{(l)}}{\sqrt{\frac{2}{B}\sum_{k=1}^{B/2}(\hat h_{kj}^{(l)}-\hat\mu_j^{(l)})^2}}\\
    &=\frac{{\vw_j^{(l)}}^T(\vh_i^{(l-1)}-\vmu^{(l-1)})}{\sqrt{\frac{2}{B}\sum_{k=1}^{B/2}({\vw^{(l)}_j}^T(\vh^{(l-1)}_i-\vmu^{(l-1)}))^2}},
\end{align}
where we denote $h_{ij}^{(l)}$, $\hat h_{ij}^{(l)}$ and $\tilde h_{ij}^{(l)}$ as the output of ReLU, convolution or fully connected layer and BN layer respectively, for $i$-th sample, $j$-th output variable at $l$-th layer.
The gradient with respect to $\vw_j^{(l)}$ is
\small
\begin{align}
    \nabla_a \vw_j^{(l)}&=\frac{\partial \gL_{a}(\tilde\vx_i)}{\partial \vw_j^{(l)}}=\frac{\nabla\tilde h_{ij}^{(l)}(\vh_i^{(l-1)}-\vmu^{(l-1)})}{\sqrt{\frac{2}{B}\sum_{k=1}^{B/2}({\vw_j^{(l)}}^T(\vh^{(l-1)}_i-\vmu^{(l-1)}))^2}}\nonumber\\
    &-\frac{\nabla\tilde h_{ij}^{(l)}{\vw^{(l)}_j}^T(\vh_i^{(l-1)}-\vmu^{(l-1)})}{(\frac{2}{B}\sum_{k=1}^{B/2}({\vw_j^{(l)}}^T(\vh^{(l-1)}_i-\vmu^{(l-1)}))^2)^{3/2}}\mSigma^{(l-1)} \vw_j^{(l)},\nonumber
\end{align}
\normalsize
where $\vmu^{(l-1)}$ and $\mSigma^{(l-1)}$ are the mean and covariance matrix from $(l-1)$-th layer, $\nabla \tilde h_{ij}^{(l)}$ represents the gradient of loss with respect to $\tilde h_{ij}^{(l)}$.
If we write the weight as its norm $\|\vw_j^{(l)}\|$ multiplied by the unit vector $\vu_j^{(l)}$, there is a relationship between $\nabla \vw_j^{(l)}$ and $\|\vw_j^{(l)}\|$,
\begin{align}
    \|\nabla_a\vw_j^{(l)}\|=\frac{1}{\|\vw_j^{(l)}\|}\|\nabla_a\vu_j^{(l)}\| \label{equ:relation}
\end{align}
\CUT{
\small
\begin{align}
    \nabla_a \gamma\vu_j^{(l)}&=\frac{\nabla\tilde h_{ij}^{(l)}(\vh_i^{(l-1)}-\vmu^{(l-1)})}{\gamma\sqrt{\frac{2}{B}\sum_{k=1}^{B/2}({\vu_j^{(l)}}^T(\vh^{(l-1)}_i-\vmu^{(l-1)}))^2}}\nonumber\\
    &-\frac{\nabla\tilde h_{ij}^{(l)}{\vu^{(l)}_j}^T(\vh_i^{(l-1)}-\vmu^{(l-1)})}{\gamma(\frac{2}{B}\sum_{k=1}^{B/2}({\vu_j^{(l)}}^T(\vh^{(l-1)}_i-\vmu^{(l-1)}))^2)^{3/2}}\mSigma^{(l-1)} \vu_j^{(l)}\nonumber\\
    &=\frac{1}{\gamma}\nabla_a \vu_j^{(l)},\\
    \Rightarrow& \|\nabla_a\vw_j^{(l)}\|=\frac{1}{\|\vw_j^{(l)}\|}\|\nabla_a\vu_j^{(l)}\|
\end{align}
\normalsize
}
This relationship has been found in \cite{hoffer2018norm,arora2018theoretical} and extended to any scale-invariant layers such as layer normalization \cite{ba2016layer} by \cite{li2020reconciling}. It means that there are two ways to increase the gradient magnitude or the effective learning rate: 1) find a steeper descent direction where $\|\nabla_a\vu_j^{(l)}\|$ is increased and 2) decrease the weight norm so that $1/\|\vw_j^{(l)}\|$ is increased. For a DNN with standard BN layers, the training will exploit this property to increase the gradient magnitude by reducing the weight norms with the help of weight decay regularization, which leads to a spurious increase in gradient magnitude and a larger variance of gradient estimation.
In contrast, for a DNN with fixed BN layers such as the Frozen Net, the gradient norm is not correlated with weight norm, 
\begin{align}
\nabla_f \vw_j^{(l)}=\frac{\partial \gL_{f}(\tilde\vx_i)}{\partial \vw_j^{(l)}}=\nabla\tilde h_{ij}^{(l)}\frac{\vh_i^{(l-1)}}{\sigma_{j,pt}^{(l)}}.
\end{align}
In this gradient, the only way to increase the gradient magnitude is to find the actual steeper direction. The overall gradient for the weight is
\begin{align}
    \Delta \vw_j^{(l)}=\nabla_a \vw_j^{(l)} + \lambda_{twins}\nabla_f \vw_j^{(l)}.
\end{align}

\paragraph{Empirical Study}
To see the difference between the gradients in AT and TWINS-AT, we record the gradient of all weight parameters in each step of the two training methods and compute the mean and STD of the gradient norm in each epoch. The model parameters and their gradients are treated as long vectors and we compute the $l_2$ norm of the weight and gradient vector. Figure~\ref{fig:grad_mean_and_std} shows the mean and normalized STD of gradient norms in 60 training epochs for three datasets. Here we show the normalized STD, i.e., dividing the STD by the mean, to see the relative effect of variance. The major finding is that the gradient magnitude of TWINS-AT is substantially larger than that of AT, while the variance of TWINS-AT is lower than AT in most epochs. Note that in theory, the ratio between STD and mean should remain the same after down-scaling the weight norm in standard BN, but in practice we do observe the high normalized variance of DNNs with standard BNs since we use one epoch's gradients to approximate the variance and mean. 

One benefit of the larger gradient magnitude is that the model can escape from the initial sub-optimal point faster and find a better local optimum than the small gradient optimization \cite{liu2020loss}. We validate this hypothesis by recording the distance between the current model and the initial model during training. Figure~\ref{fig:weight_dist_compare} shows these weight distances  on three datasets, where TWINS-AT moves away from the initial model much faster than the AT baseline. The robust overfitting effect of adversarial training \cite{rice2020overfitting} is partially a result of large gradient variance, especially at the final stage of training \cite{chen2020robust}. We compare the robust accuracy drop of AT and TWINS-AT in Table~\ref{tab:robust_overfitting} and find that the small relative variance of TWINS-AT has the effect of reducing robust overfitting. See the experiment details in Sec.~\ref{sec:experiment} and Supp.


\begin{table*}[t]
\centering
\small
\begin{tabular}{|c|l|l|l|l|l|l|}
\hline
        Metric          &     Method     & CIFAR10 & CIFAR100 & Caltech256 & CUB200 & Stanford Dogs \\ \hline
\multirow{6}{*}{Clean Acc.} & AT              &  89.77 & 69.48         & 75.90 & 65.74 & 60.09 \\  
                           & TWINS-AT     & 91.24(\increase{+1.47})  &  70.72(\increase{+1.24})  &   76.86(\increase{+0.96}) & 68.09(\increase{+2.35}) & 64.98(\increase{+4.89})  \\ 
                           & TWINS-AT+warmup &  91.95(\increase{+2.18})  &    72.12(\increase{+2.64})      &       77.35(\increase{+1.45})     &  67.64(\increase{+1.90})   &  66.12(\increase{+6.03}) \\ \cline{2-7}
                           & TRADES              &   87.06  & 62.76 & 69.70 & 58.92  & 59.99 \\  
                           & TWINS-TRADES        & 86.61(\decrease{-0.45})   & 66.72(\increase{+3.96})  & 71.12(\increase{+1.42})  &  60.72(\increase{+1.80})    &  60.58(\increase{+0.59})    \\ 
                           & TWINS-TRADES+warmup &  86.60(\decrease{-0.46})   &   65.91(\increase{+3.15})    & 73.39(\increase{+3.69})         &  61.05(\increase{+2.13})   &  63.96(\increase{+3.97}) \\ \hline
\multirow{6}{*}{PGD10}       & AT              &   52.24   & 28.52  &  48.37   &   26.60  & 19.80  \\ 
                           & TWINS-AT        &  52.73(\increase{+0.49})   &   31.08(\increase{+2.56})      &   48.40(\increase{+0.03})   &  29.24(\increase{+2.64})   & 20.58(\increase{+0.78})   \\  
                           & TWINS-AT+warmup &    52.46(\increase{+0.22})    &      29.12(\increase{+0.60})     &         49.13(\increase{+0.76})   &   27.67(\increase{+1.07})  &    19.48(\decrease{-0.32})   \\ \cline{2-7}
                           & TRADES              & 54.04  & 32.20& 47.28 & 27.87  & 21.36 \\ 
                           & TWINS-TRADES        & 56.23(\increase{+2.19}) &  33.51(\increase{+1.31}) &  47.31(\increase{+0.03})  & 27.05(\decrease{-0.82})  & 19.93(\decrease{-1.43})  \\ 
                           & TWINS-TRADES+warmup & 55.81(\increase{+1.77})   &   33.48(\increase{+1.28})   &  48.53(\increase{+1.25})        &  26.68(\decrease{-1.19})   &   19.46(\decrease{-1.90})   \\ \hline
\multirow{6}{*}{AA}        & AT              &  48.46  &  23.47 &  43.85    &  22.82 &  12.30 \\ 
                           & TWINS-AT        & \textbf{49.81}(\increase{+1.35})  & \textbf{26.73}(\increase{+3.26})  &   43.69(\decrease{-0.16})  &  22.33(\decrease{-0.49})  & \textbf{14.37}(\increase{+2.07})  \\ 
                           & TWINS-AT+warmup &  49.02(\increase{+0.56})   &  25.72(\increase{+2.25})        &         \textbf{43.92}(\increase{+0.07})   &   \textbf{23.58}(\increase{+0.75})  &   13.80(\increase{+1.50})    \\ \cline{2-7}
                           & TRADES              &  50.31   &   26.40   &  43.39  &  22.21   & 12.05   \\  
                           & TWINS-TRADES        &  \textbf{51.71}(\increase{+1.40})   &    28.29(\increase{+1.89})    &  41.77(\decrease{-1.62})         &   \textbf{22.68}(\increase{+0.47})  &  \textbf{13.36}(\increase{+1.31})  \\ 
                           & TWINS-TRADES+warmup &  51.10(\increase{+0.79})   &  \textbf{28.30}(\increase{+1.90})    & \textbf{43.55}(\increase{+0.16})        &   21.92(\decrease{-0.29})  &    10.94(\decrease{-1.11})   \\ \hline
\end{tabular}
\vspace{-0.3cm}
\caption{The performance of our TWINS-AT and TWINS-TRADES on five image classification tasks compared with AT and TRADES. The clean accuracy means the accuracy when testing images are input without adversarial perturbations. PGD10 and AA denote the robust test accuracy under PGD10 and AutoAttack. The increase and decrease in performance are denoted with \increase{green} and \decrease{red} numbers. The \textbf{bold} numbers denote the best robust accuracy under AA. The proposed TWINS achieves better robustness and clean accuracy compared with the baseline. Averaged over the datasets, the clean and robust accuracy of TWINS are increased by 2.18\% and 1.21\% compared with AT, and 1.46\% and 0.69\% compared with TRADES. The means and STDs of the performance are in the supplemental. }
\vspace{-0.5cm}
\label{tab:main_table}
\end{table*}

\begin{table}[]
\small
\begin{tabular}{|l|l|c|c|c|}
\hline
Metric                     & PT Model       & Caltech256 & CUB200 & Dogs \\ \hline
\multirow{3}{*}{Clean Acc} & Random    &  48.99   &   12.38   &     7.27     \\ 
                           & Non-Robust &  64.66   &   53.30   &    41.99     \\ 
                           & Robust     &   \textbf{75.90}  &   \textbf{65.74}   & \textbf{60.09}  \\ \hline
\multirow{3}{*}{PGD10}     & Random     &  31.78   &  3.728    &    3.59     \\ 
                           & Non-Robust &  39.86   &   19.68  & 13.32    \\  
                           & Robust     &   \textbf{48.37}  &   \textbf{26.60}   &   \textbf{19.80}  \\ \hline
\end{tabular}
\vspace{-0.3cm}
\caption{\ziquan{Comparison of random initialization, non-robust and robust pre-trained model on three difficult classification tasks, when fine-tuned with AT. The robust pre-trained model is indispensable to downstream robustness. }}
\vspace{-0.7cm}
\label{tab:important_robust_pt}
\end{table}

\section{Experiment}
\label{sec:experiment}
This section presents our experiment with TWINS. We first introduce our experiment setting and then show our main result and ablation study.

\subsection{Experiment Settings}
\vspace{-0.25cm}
\noindent\textbf{Dataset.} We use five datasets in our experiment. CIFAR10 and CIFAR100 \cite{krizhevsky2009learning} are low-resolution image datasets, where the training and validation sets have 50,000 and 10,000 images, and CIFAR10 has 10 classes, while CIFAR100 has 100 classes. Caltech-256 \cite{griffin2007caltech} is a high-resolution dataset with 30,607 images and 257 classes, which is split into training and validation set with a ratio of 9:1. Caltech-UCSD Birds-200-2011 (CUB200) \cite{WahCUB_200_2011} is a high-resolution bird image dataset for fine-grained image classification, which contains 200 classes of birds, 5,994 training images and 5,794 validation images. Stanford Dogs \cite{khosla2011novel} has high-resolution dog images from 120 dog categories, where the training and validation set has 12,000 and 8,580 images. For both low-resolution image datasets (CIFAR10 and CIFAR100) and high-resolution datasets (Caltech-256,CUB200 and Stanford Dogs), we resize the image to 224$\times$224 so that the input sizes are the same for pre-training and fine-tuning. As with pre-training, the input image is normalized by the mean and STD of the pre-training set. Note that the resizing and normalization function is integrated into the model so we can attack the input image with the [0,1] bounds for pixel values as in standard adversarial attacks. We use the standard ImageNet data augmentation for high-resolution datasets \cite{he2016deep}. For CIFAR datasets, we use random cropping with padding=4 and random horizontal flipping.

\noindent\textbf{Adversarial Pre-Training.} Large-scale adversarial pre-training on ImageNet is time-consuming, and thus \cite{salman2020adversarially} has released adversarially pre-trained ResNet50 and WideResNet50-2, trained with $l_2$ and $l_{\infty}$ norm bounded attacks. 
In this paper, we adopt the pre-trained ResNet50 models,  trained with $l_{\infty}$  attack with bound $\epsilon_{pt}=4/255$. We test other robust pre-trained models in our ablation study.

\noindent\textbf{Training Setting.} For baselines and our method, we train all parameters of the pre-trained model, i.e., \emph{full fine-tuning} instead of linear probing \cite{shafahi2020adversarially}, with PGD attacks of $l_{\infty}$ norm. The PGD step is 10, $\epsilon_{ft}=8/255$ and stepsize $\alpha=2/255$. We set the batch size as 128 and train the model for 60 epochs and divide the learning rate by 0.1 at 30th and 50th epoch. In TWINS with warmup, we initialize the means and STDs with their pre-trained values, and update the means and STDs using the target training set. The momentum of updating statistics is 0.1, the batch size is 128, and the warmup only lasts one epoch. Note that in the warmup stage, the input samples are added with adversarial perturbations generated by the PGD attack, which has the same setting as the attack in training, and the classifier layer is the pre-trained classifier for the adversarial attack. Our pilot experiment shows that using adversarial examples as input is more effective than using clean exmaples in the warmup. The optimizer is SGD with momentum in all of our experiments. The learning rate, weight decay and regularization hyperparameter are determined by grid search, which is described in detail in the supplemental. 

\noindent\textbf{Adversarial robustness evaluation.} Two standard adversarial attacks are used in our experiment, i.e., PGD and AutoAttack \cite{croce2020reliable}. The adversarial perturbation is $l_{\infty}$ norm bounded in our evaluation. The setting of PGD attack for validation set is the same as the PGD attack in training. The AutoAttack (AA) is a more reliable adversarial attack and more often used \abc{for evaluation} than PGD in recent years. We use the standard attacks of AA, i.e., untargeted APGD-CE, targeted APGD-DLR, targeted FAB \cite{croce2020minimally} and Square Attack \cite{andriushchenko2020square}, 
with $\epsilon=8/255$.
The robust accuracy in our experiment result denotes the accuracy under AA. 

\begin{table}[t]
\begin{center}
\small
\begin{tabular}{|c|l|l|l|l|l|}
\hline
 \multirow{2}{*}{Method}& \multirow{2}{*}{Metric}
 &\multicolumn{4}{c|}{$\lambda_{WD}$}\\
 \cline{3-6}
 & &
  1e-5 &
  1e-4 &
  1e-3 &
  1e-2 \\ \hline
\multirow{2}{*}{AT}       & Clean Acc.  & 89.92 & 89.94 & 89.77 & \textbf{90.28}  \\  
                          & Robust Acc.  & 46.34 & 44.67 & 48.46 & 47.57    \\ 
                 \hline
\multirow{2}{*}{TWINS-AT}       & Clean Acc.  & \textbf{91.90} & \textbf{91.42} & \textbf{91.24} & 87.33  \\  
                          & Robust Acc.  & \textbf{47.19} & \textbf{49.07} & \textbf{49.81} & \textbf{51.65}    \\ 
 \hline
\end{tabular}
\vspace{-0.3cm}
\caption{The performance of TWINS-AT and AT on CIFAR10 when the hyperparameter for weight decay $\lambda_{WD}$ is changed. The robust accuracy is evaluated using AutoAttack. Our TWINS-AT achieves better adversarial robustness than AT for different $\lambda_{WD}$.}
\vspace{-0.7cm}
\label{tab:weight_decay}
\end{center}
\end{table}

\subsection{Experimental Result}
On the three high-resolution datasets, we compare the performance of fine-tuning different initialization models, i.e., random initialization, standard pre-trained ResNet50 and robust pre-trained ResNet50. Table~\ref{tab:important_robust_pt} shows that the pre-trained models are essential to the accuracy and robustness in challenging downstream tasks, since the random initialization is much worse than the two pre-trained models. The robust pre-trained model has a clear benefit over the standard pre-trained one, indicating that robust pre-training is indispensable to downstream robustness.

Table~\ref{tab:main_table} shows the result of TWINS-AT and TWINS-TRADES compared with the baselines. Since AA is a more reliable attack, we highlight the best robust accuracy under AA on each dataset. The TWINS fine-tuning learns more robust DNNs, as well as achieves a better clean accuracy on all five datasets, demonstrating the strong effectiveness of TWINS in the robust pre-training and robust fine-tuning setting. On CIFAR10 and CIFAR100, both TWINS-AT and TWINS-TRADES achieve better robustness and clean accuracy than their baselines, while the warmup only improves the clean accuracy and hurts the robustness sometimes. On Caltech-256,  TWINS-AT improves upon the vanilla AT in both robustness and accuracy, but TWINS-TRADES does not perform better than vanilla TRADES. However, the warmup helps boost the performance of TWINS-TRADES as well as TWINS-AT and makes the TWINS with warmup perform better than the baselines. 

On the two fine-grained image classification datasets, TWINS-AT and TWINS-TRADES generally perform better than baselines in terms of accuracy and robustness, if we look at the robust accuracy under AA, where only TWINS-AT on CUB200 has a slightly worse robust accuracy than its baseline. We find that the warmup improves the clean accuracy but hurts the robustness in most cases, except for CIFAR100 and Caltech-256. This can be a result of noisy adversarial perturbation, since we use the pre-trained classifier layer in the adversarial attack, or the insufficient update steps. Nevertheless, we note that the warmup can be considered as an operation for achieving a trade-off between robustness and accuracy.

\begin{figure}[t]
    \centering
    \includegraphics[width=1.0\linewidth]{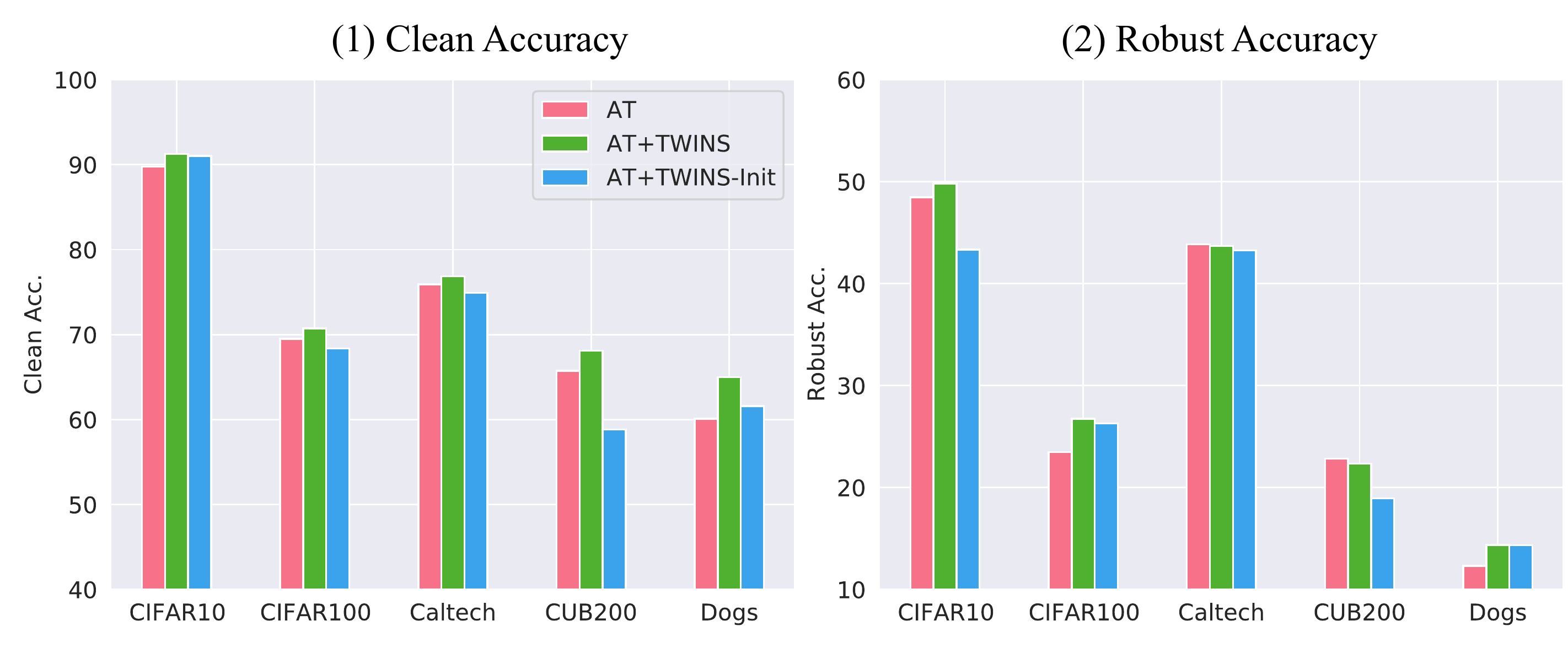}
    \vspace{-0.95cm}
    \caption{Ablation study where the BN statistics in TWINS are initialized with (0,1) for means and STDs, denoted as TWINS-Init. The population mean and STD of pre-training are crucial to TWINS. 
    }
    \vspace{-0.6cm}
    \label{fig:ablation_frozen_bn_no_pretrain}
\end{figure}

\subsection{Ablation Study}


\noindent\textbf{TWINS initialization.} We use the pre-trained BN statistics in the Frozen Net to keep the robust information learned during pre-training. To show the importance of the pre-trained BN statistics, we use the standard initialization (mean=0 and STD=1) for BN statistics in the Frozen Net, \abc{denoted as TWINS-Init}, and show the result on the five datasets in Figure~\ref{fig:ablation_frozen_bn_no_pretrain}. Both clean and robust accuracy drop when the (0,1) initialization is used in TWINS-Init, demonstrating the crucial role of pre-trained statistics in TWINS. The fact that TWINS-Init sometimes improves upon the AT baseline motivates us to investigate the effect of TWINS on gradient norms in Section~\ref{sec:twins}.

\noindent\textbf{Effect of weight decay.} Weight decay is the reason for the decreasing weight norm in DNNs with BN layers, so increasing the hyperparameter of weight decay $\lambda_{WD}$ is also a way to increase the gradient magnitude. Table~\ref{tab:weight_decay} show the result of TWINS-AT and AT when different $\lambda_{WD}$ are used when fine-tuning the robust pre-trained model on CIFAR10. The robust accuracy of TWINS-AT is consistently better than that of AT across different $\lambda_{WD}$'s. We draw the same conclusion on CIFAR100 (see supplemental). Note that the clean accuracy of TWINS-AT drops when a large $\lambda_{WD}$ is used, suggesting that we should not use a too large $\lambda_{WD}$ for TWINS-AT. 

\noindent\textbf{Different robust pre-trained models.} The main experiment uses the robust pre-trained ResNet50 with $\epsilon_{pt}=4/255$ as the initial model. We try different robust pre-trained models with different $\epsilon_{pt}$ in Table~\ref{tab:pretrain_eps}, which shows that a larger $\epsilon_{pt}$ is beneficial to both clean and robust accuracy in the downstream, and the proposed TWINS-AT is better than AT in both metrics with different pre-trained models. 

\begin{table}[t]
\begin{center}
\small
\begin{tabular}{|c|l|l|l|l|l|}
\hline
 \multirow{2}{*}{Method}& \multirow{2}{*}{Metric}
 &\multicolumn{4}{c|}{$\epsilon_{pt}$}\\
 \cline{3-6}
 & &
  1/255 &
  2/255 &
  4/255 &
  8/255 \\ \hline
\multirow{2}{*}{AT}       & Clean Acc.  & 65.47 & 67.08 & 69.48 & 69.93  \\  
                          & Robust Acc.  & 24.76 & 25.79 & 23.47 & 27.71    \\ 
                 \hline
\multirow{2}{*}{TWINS-AT}       & Clean Acc.  & \textbf{68.55} & \textbf{70.45} & \textbf{70.72} & \textbf{72.59}  \\  
                          & Robust Acc.  & \textbf{25.97} & \textbf{26.62} & \textbf{26.73} & \textbf{28.62}    \\ 
 \hline
\end{tabular}
\vspace{-0.3cm}
\caption{The performance of TWINS-AT and AT on CIFAR100 when robust pre-trained ResNet50 with different $\epsilon_{pt}$ are used. 
}
\vspace{-0.8cm}
\label{tab:pretrain_eps}
\end{center}
\end{table}

\vspace{-0.2cm}
\section{Conclusion}
\label{sec:conclusion}
\vspace{-0.3cm}
This paper investigates the utility of robust pre-trained models in various downstream classification tasks. We first find that the commonly used data- and model-based approaches to maintain pre-training information do not work in the adversarial robust fine-tuning. We then propose a subtle statistics-based method, TWINS, for retaining the pre-training robustness in the downstream. In addition to the robustness preserving effect, we find that TWINS increases the gradient magnitude without sacrificing the training stability and improves the training dynamics of AT. Finally, the performance of TWINS is shown to be stronger than that of AT and TRADES on five datasets.
One limitation of our work is that we only evaluate the robust supervised pre-trained ResNet50. Recently, robust pre-trained ViT's on ImageNet \cite{bai2021transformers} have been released. Our statistics-based approach can be extended to the layer normalization, on which the increasing gradient magnitude argument also holds, and thus future work will extend TWINS to ViT.

\noindent\textbf{Acknowledgement} This work was supported by a grant from
the Research Grants Council of the Hong Kong Special Administrative Region, China (Project No. CityU 11215820) and the Fundamental Research Funds for the Central University of China (DUT No. 82232031).

{\small
\bibliographystyle{ieee_fullname}
\bibliography{egbib}
}

\clearpage
\begin{table*}[h]
\centering
\begin{tabular}{|l|l|l|l|l|l|}
\hline
               Method     & CIFAR10 & CIFAR100 & Caltech256 & CUB200 & Stanford Dogs \\ \hline
 AT              &  (1e-3,1e-3) & (1e-3,1e-4)         & (3e-3,1e-3) & (1e-2,1e-4) & (1e-3,1e-4) \\  
                            TWINS-AT     & (3e-3,1e-3,1.0)  &  (1e-3,1e-4,0.3)  &   (3e-3,1e-3,0.4) & (3e-3,1e-4,0.3) & (3e-3,1e-4,1.0)  \\ 
                            TWINS-AT+warmup &  (3e-3,1e-3,1.0)  &    (1e-3,1e-4,1.0)      &       (3e-3,1e-3,0.4)      &  (1e-2,1e-4,0.3)   &  (3e-3,1e-4,0.3) \\ \cline{1-6}
                            TRADES              &   (1e-2,1e-4)  & (1e-3,1e-4) & (1e-2,1e-4) & (1e-2,1e-4)  & (1e-3,1e-4) \\  
                            TWINS-TRADES        & (1e-2,1e-4,1.0)   & (1e-2,1e-4,1.0)  & (3e-3,1e-4,1.0)  &  (1e-2,1e-4,3.0)   &  (3e-3,1e-4,1.0)    \\ 
                           TWINS-TRADES+warmup &  (1e-2,1e-4,1.0)   &   (1e-2,1e-4,1.0)    & (3e-3,1e-4,0.4)         &  (1e-2,1e-4,0.3)   &  (1e-3,1e-4,1.0) \\ \hline
\end{tabular}
\caption{The hyperparameters of AT, TRADES and TWINS in our experiment. The format means ($\eta$, $\lambda_{WD}$, $\lambda_{twins}$) for TWINS and ($\eta$, $\lambda_{WD}$) for baselines.}
\label{tab:main_table_hyper}
\end{table*}

\section*{A. Experimental Details}
Fig.~\ref{fig:compare_data_and_model_approach} shows the experiment result of UOT fine-tuning and Learning without Forgetting. In the UOT data selection, we follow the same experiment setting as in \cite{liu2022improved}, i.e., the distance function in UOT is the cosine-based distance and $\epsilon_c=0.01$. In all the experiments, we search the hyperparameters using grid search and determine the optimal hyperparameter based on the performance of the validation set. In UOT fine-tuning, $\lambda_{UOT}$ is searched from $\{0.001,0.01,0.1,1.0,10.0\}$, learning rate is searched from $\{0.001,0.003,0.01\}$ and weight decay is searched from $\{$1e-5,1e-4,1e-3$\}$. In LwF, learning rate and weight decay are searched with the same range as in UOT fine-tuning and $\lambda_{LwF}$ is searched from $\{$1e-5,1e-4,1e-3,1e-2,1e-1$\}$. In Fig.~\ref{fig:compare_data_and_model_approach}a, UOT's hyperparameter is $\lambda_{UOT}=0.01,\eta=0.001,\lambda_{WD}=\text{1e-4}$ and LwF's hyperparameter is $\lambda_{LwF}=\text{1e-4},\eta=0.001,\lambda_{WD}=\text{1e-4}$. In Fig.~\ref{fig:compare_data_and_model_approach}b, UOT's hyperparameter is $\lambda_{UOT}=0.01,\eta=0.01,\lambda_{WD}=\text{1e-4}$ and LwF's hyperparameter is $\lambda_{LwF}=\text{1e-2},\eta=0.003,\lambda_{WD}=\text{1e-4}$.

In Tab.~\ref{tab:main_table_hyper}, we list the hyperparameters of baselines and TWINS in Tab.~\ref{tab:main_table}. The $\beta$ parameter in TRADES and TWINS-TRADES is fixed as 6.0, which is the default value in the original TRADES experiment \cite{zhang2019theoretically}, so we only search the learning rate and weight decay parameter in TRADES. The search range for learning rate $\eta$ is \{3e-4,1e-3,3e-3,1e-2,3e-2\}, for weight decay $\lambda_{WD}$ is \{1e-5,1e-4,1e-3,1e-2\}, for $\lambda_{twins}$ is \{0.1,0.2,0.3,0.4,0.5,1.0\}. 

Tab.~\ref{tab:robust_overfitting} shows the best and final robust accuracy under PGD10 attack. The robust accuracy is not evaluated using AutoAttack for the efficiency since we evaluate the model for every epoch. To save space, we use C10 and C100 to denote CIFAR10 and CIFAR100. Caltech, CUB and Dogs are short for Caltech-256, CUB200 and Stanford Dogs. In Tab.~\ref{tab:important_robust_pt}, we use the same hyperaparameter search for random initialization and standard pre-trained model initialization. For random initialization, we increase the training time to 120 epochs and decay the learning rate at 60 and 100 epochs to improve the performance and make a more fair comparison. Tab.~\ref{tab:weight_decay} and Tab.~\ref{tab:weight_decay_cifar100} show the performance of AT and TWINS-AT on CIFAR10 and CIFAR100 when $\lambda_{WD}$ is changed. We keep other hyperparameters untouched and only change the $\lambda_{WD}$. Tab.~\ref{tab:pretrain_eps} changes the pre-trained model and keeps other hyperparameters the same as in Tab.~\ref{tab:main_table_hyper}.

For AT and TWINS-AT, we run the experiment on Nvidia-3090-24G GPU. For TRADES and TWINS-TRADES, we run the experiment on Nvidia-V100-32G GPU since they require a double batch size in effect. We run each experiment twice for baselines and TWINS and report the one with the optimal robustness performance.

Algorithm \ref{alg:twins} gives the pseudo-code  of TWINS-AT and TWINS-TRADES for reference. We include our code for TWINS-AT and TWINS-TRADES in the supplemental and encourage the readers to run our code.

\section*{B. Derivation of Equation (\ref{equ:relation})}
We include the derivation of Equation (\ref{equ:relation}) to make our paper self-contained. We use the $\gamma$ to represent the weight norm $\|\vw_j^{(l)}\|$, then the gradient of $\vw_j^{(l)}$ in Adaptive Net is written as $\nabla_a\gamma\vu_j^{(l)}$ and we extract the norm term $\gamma$ out of the gradient,
\small
\begin{align}
    \nabla_a \gamma\vu_j^{(l)}&=\frac{\nabla\tilde h_{ij}^{(l)}(\vh_i^{(l-1)}-\vmu^{(l-1)})}{\gamma\sqrt{\frac{2}{B}\sum_{k=1}^{B/2}({\vu_j^{(l)}}^T(\vh^{(l-1)}_i-\vmu^{(l-1)}))^2}}\nonumber\\
    &-\frac{\nabla\tilde h_{ij}^{(l)}{\vu^{(l)}_j}^T(\vh_i^{(l-1)}-\vmu^{(l-1)})}{\gamma(\frac{2}{B}\sum_{k=1}^{B/2}({\vu_j^{(l)}}^T(\vh^{(l-1)}_i-\vmu^{(l-1)}))^2)^{3/2}}\mSigma^{(l-1)} \vu_j^{(l)}\nonumber\\
    &=\frac{1}{\gamma}\nabla_a \vu_j^{(l)},\\
    \Rightarrow& \|\nabla_a\vw_j^{(l)}\|=\frac{1}{\|\vw_j^{(l)}\|}\|\nabla_a\vu_j^{(l)}\|
\end{align}
\normalsize
This gives us the relationship between weight norm and its gradient norm.

\section*{C. More Experiment Result}

Tab.~\ref{tab:weight_decay_cifar100} shows the performance of TWINS-AT when different $\lambda_{WD}$'s are used. For $\lambda_{WD}$=1e-4, 1e-3 and 1e-2, the adversarial robustness of TWINS-AT is better than the AT baseline. Note that as in the CIFAR10 experiment, TWINS-AT does not achieve a better robust accuracy when $\lambda_{WD}$ is too large, i.e., 1e-2. On other 3 WD settings, the clean accuracy of TWINS is clearly better than AT. Thus, the performance of TWINS-AT is not very sensitive to the weight decay hyperparameter if we use $\lambda_WD$=1e-3 or 1e-4.

\begin{figure*}[t]
    \centering
    \includegraphics[width=0.65\linewidth]{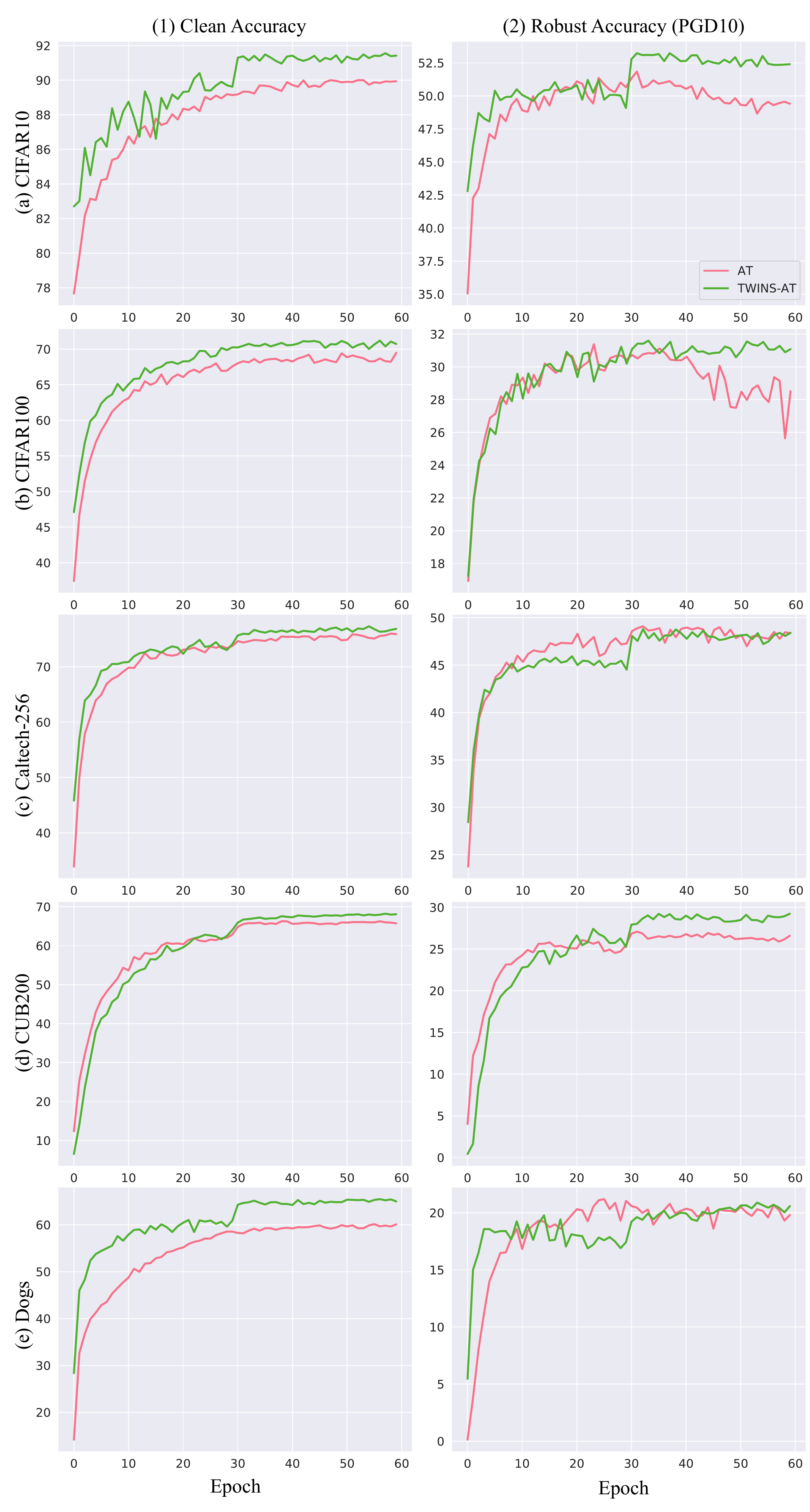}
    \caption{The curve of clean accuracy and robust accuracy (PGD10 attack) on validation set.
    }
    \label{fig:learning_curve}
\end{figure*}

The clean and robust accuracy on the validation set in each epoch of AT and TWINS-AT are shown in Fig.~\ref{fig:learning_curve}. On CIFAR10, CIFAR100, Caltech, and Dogs, the convergence of TWINS-AT is faster than the AT baseline, as a result of larger effective learning rate and faster escape from initialization. The figure also shows that the robust overfitting effect is obvious on CIFAR10 (Fig.~\ref{fig:learning_curve}.a2) and CIFAR100 (Fig.~\ref{fig:learning_curve}.b2), while TWINS-AT is less prone to the robust overfitting. 

Table \ref{tab:std_result} shows the experiment result of running three trials for AT and TWINS-AT. The improvement of TWINS over AT is significant in most cases. 

We run an experiment using the DTD dataset \cite{cimpoi14describing}, which only contains texture images and looks very different from ImageNet. The clean (robust) accuracy of TWINS-AT and AT is 57.39\% (21.54\%) and 55.96\% (21.17\%). Similar to \cite{liu2020improve}, we find that although the domain difference is large, keeping the pre-training information during  fine-tuning is beneficial, especially for the clean accuracy. 

We ran the experiment of only fine-tuning the classification layer \abc{and updating the BN stats, while fixing the backbone weights.} Table \ref{tab:robust_feature} shows that learning with the pre-trained robust features achieves high clean accuracy but low robust accuracy, consistent with our experiment result that the warmup in TWINS mainly benefits the clean accuracy.

We try different $\lambda_{twins}$ on CIFAR10 in Table \ref{tab:hyperpar}. TWINS is not very sensitive to $\lambda_{twins}$, when it is not very large.
\begin{table}[h!]
\centering
\scriptsize
\begin{tabular}{|l|l|l|l|l|l|}
\hline
           & 1e-1 & 3e-1 & 1.0 & 3.0 & 10.0 \\ \hline
Clean Acc. & 91.18  & 91.56    & 91.24       & 90.46   & 10.0    \\
AA         & 49.11  & 48.51    & 49.8       & 48.51   & 10.0  \\ \hline
\end{tabular}
\caption{Performance of TWINS with different $\lambda_{twins}$ on CIFAR10.}
\label{tab:hyperpar}
\end{table}

\begin{table}[h!]
\centering
\scriptsize
\begin{tabular}{|c|c|c|c|c|c|c|}
\hline
&\multicolumn{3}{c|}{\abc{Fine-tune} Adv. Train}&\multicolumn{3}{c|}{\abc{Fine-tune} Std. Train}\\
\hline
Dataset & C10 & C100 & Caltech &C10 & C100 & Caltech \\
\hline
Clean Acc. &79.48 & 63.28 & 78.97 & 92.11 & 77.92 & 81.41\\
PGD10 & 12.46 & 9.84 & 36.17 & 0.67 & 1.00 & 21.01 \\
\hline
\end{tabular}
\caption{Adversarial and standard fine-tuning with fixed backbone weights and adaptive BN stats.}
\label{tab:robust_feature}
\end{table}

\begin{table}[h]
\tiny
\begin{tabular}{|l|l|l|l|l|l|l|}
\hline
         & Metric  & C10 & C100 & Caltech & CUB & Dogs \\ \hline
\multirow{2}{*}{AT} & Clean & 89.81(0.16)   & 69.02(0.42)   & 75.31(0.55)      & 65.51(0.25)  & 59.50(0.54)  \\
& Rob. & 47.56(1.15)  & 24.93(1.29)   & \textbf{43.39(0.52)}    & \textbf{23.25(0.39)}  & 11.90(0.68) \\
\hline
\multirow{2}{*}{TWINS-AT} & Clean & \textbf{91.56(0.38)}  & \textbf{70.95(0.55)}   & \textbf{76.92(0.50)}     & \textbf{67.89(0.17)}   & \textbf{65.25(0.26)}   \\
& Rob. & \textbf{48.99(0.89)} & \textbf{26.06(0.62)}   & 42.43(1.09) & 21.89(0.38) & \textbf{13.95(0.54)} \\
\hline
\end{tabular}
\caption{Mean and std of AT and TWINS.}
\label{tab:std_result}
\vspace{-0.4cm}
\end{table}

\begin{table}[H]
\begin{center}
\small
\begin{tabular}{|c|l|l|l|l|l|}
\hline
 \multirow{2}{*}{Method}& \multirow{2}{*}{Metric}
 &\multicolumn{4}{c|}{$\lambda_{WD}$}\\
 \cline{3-6}
 & &
  1e-5 &
  1e-4 &
  1e-3 &
  1e-2 \\ \hline
\multirow{2}{*}{AT}       & Clean Acc.  & 68.51 & 69.48 & 67.92 & \textbf{66.48}  \\  
                          & Robust Acc.  & \textbf{25.45} & 23.47 & \textbf{26.69} & 27.16    \\ 
                 \hline
\multirow{2}{*}{TWINS-AT}       & Clean Acc.  & \textbf{71.64} & \textbf{70.72} & \textbf{70.74} & 65.74  \\  
                          & Robust Acc.  & 24.65 & \textbf{26.73} & \textbf{26.69} & \textbf{27.81}    \\ 
 \hline
\end{tabular}
\caption{The performance of TWINS-AT and AT on CIFAR100 when the hyperparameter for weight decay $\lambda_{WD}$ is changed.}
\label{tab:weight_decay_cifar100}
\end{center}
\end{table}

\CUT{
We try the mixed BN that combines the pre-training statistics and target statistics with weight $\lambda_{mix}$, i.e.,
\begin{align}
    \mu_{mix}&=(1-\lambda_{mix})\mu_{pt}+\lambda_{mix}\mu,\\
    \sigma_{mix}^2&=(1-\lambda_{mix})\sigma_{pt}^2+\lambda_{mix}\sigma^2,
\end{align}
where the $\mu$ and $\sigma^2$ are current batch's mean and variance in training and population mean and variance in inference. Using the mixed BN, the relationship between weight norm $\|\vw_j^{(l)}\|$ and its gradient norm $\|\nabla \vw_j^{(l)}\|$ is also broken, since the batch normalization is no longer scale-invariant. Thus, AT with this mixed BN strucuture should also perform better than the vanilla AT baseline. We run the experiment on CIFAR10 and determine the hyperparameters $\lambda_{mix}$ from $\{0.1,0.5,0.9\}$ and learning rate $\eta$ from $\{0.001,0.003\}$ using grid search. Tab.~\ref{tab:MBN} shows the performance of AT with Mixed BN (MBN-AT) and compares with AT and TWINS-AT. It turns out that the mixed BN learns more robust DNNs than standard fine-tuning with standard BN structure, which validates our argument that the benefit of training dynamics is important. Compared with TWINS-AT, the MBN-AT fails to improve the clean accuracy, since the population statistics are not fully kept but mixed with downstream statistics.} 

\CUT{
\begin{table}[H]
\centering
\begin{tabular}{|l|l|l|l|}
\hline
            & AT    & TWINS-AT & MBN-AT \\ \hline
Clean Acc.  & 89.77 & \textbf{91.24}    & 88.95  \\ \hline
Robust Acc. & 48.46 & 49.81    & \textbf{50.62}  \\ \hline
\end{tabular}
\caption{The performance of standard AT, TWINS-AT and MBN-AT on CIFAR10. }
\label{tab:MBN}
\end{table}
}

\begin{algorithm*}[t]
 \caption{Training algorithm of TWINS-AT and TWINS-TRADES}
  \KwIn{Training data $\gD_{tr}$, TWINS parameter $\lambda_{twins}$, learning rate $\eta$, parameter of TRADES $\beta$, number of epochs $N_e$, batch size $B$ }
  \KwOut{Model parameters ($\vtheta$,$\vw$,$b$)}
  Initialize model parameters and BN statistics with pre-trained parameters and BN statistics;
  \\
    \For{$i=1,\dots,N_e$}
    {
      Adjust $\eta$;\\
      Split $\gD_{tr}$ into $N_B=$ceil$(N_{tr}/B)$ batches;
      \\
      \For{$b=1,\dots,N_B$}
      {
        Generate adversarial examples $\{\tilde\vx_i^{(a)},\vy_i\}_{i=1}^{B}$ by attacking Adaptive Net $\vtheta_a$;\\
        \If{\textcolor{orange}{AT}} {\textcolor{orange}{$Loss= \sum_{j=1}^{B/2}\gL(\vw^T g_{\vtheta_{a}}(\tilde{\vx}_j^{(a)})+b,\vy_j) +\lambda_{twins}\sum_{i=B/2+1}^{B}\gL(\vw^T g_{\vtheta_{f}}(\tilde{\vx}_i^{(a)})+b,\vy_i$);}}
        \If{\textcolor{blue}{TRADES}} {\small\textcolor{blue}{$Loss=\frac{2}{B}\sum_{j=1}^{B/2}\gL(\vw^T g_{\vtheta_{a}}(\tilde{\vx}_j^{(a)})+b,\vy_j)+\beta\frac{2}{B}\sum_{j=1}^{B/2}D_{KL}(\vw^Tg_{\vtheta_a}(\tilde\vx_j^{(a)})+b,\vw^Tg_{\vtheta_a}(\vx_j)+b)+$}\\
        \textcolor{blue}{$\quad\lambda_{twins}[\frac{2}{B}\sum_{i=B/2+1}^{B}\gL(\vw^T g_{\vtheta_{f}}(\tilde{\vx}_i^{(a)})+b,\vy_i)+\beta\frac{2}{B}\sum_{i=B/2+1}^{B}D_{KL}(\vw^Tg_{\vtheta_f}(\tilde\vx_i^{(a)})+b,\vw^Tg_{\vtheta_f}(\vx_i)+b)]$;}}
        {$(\vtheta,\vw,b) := (\vtheta,\vw,b) - \eta\nabla_{(\vtheta,\vw,b)}Loss\quad$ \textcolor{gray}{\# $\vtheta_a$ and $\vtheta_f$ share the parameter $\vtheta$}}
      }
    }
    \label{alg:twins}
\end{algorithm*}

\end{document}

%% file: math_commands.tex
\usepackage{amsmath,amsfonts,bm}

\def\eqref#1{equation~\ref{#1}}

\def\1{\bm{1}}

\def\vmu{{\bm{\mu}}}
\def\vtheta{{\bm{\theta}}}

\def\vh{{\bm{h}}}

\def\vu{{\bm{u}}}

\def\vw{{\bm{w}}}
\def\vx{{\bm{x}}}
\def\vy{{\bm{y}}}

\def\mSigma{{\bm{\Sigma}}}

\DeclareMathAlphabet{\mathsfit}{\encodingdefault}{\sfdefault}{m}{sl}
\SetMathAlphabet{\mathsfit}{bold}{\encodingdefault}{\sfdefault}{bx}{n}

\def\gD{{\mathcal{D}}}

\def\gL{{\mathcal{L}}}

\def\sP{{\mathbb{P}}}
\def\sQ{{\mathbb{Q}}}

